\documentclass[lettersize,journal]{IEEEtran}
\usepackage{amsmath,amsfonts}
\usepackage{algorithmic}
\usepackage{algorithm}
\usepackage{array}
\usepackage[caption=false,font=normalsize,labelfont=sf,textfont=sf]{subfig}
\usepackage{textcomp}
\usepackage{booktabs} % for professional tables
\usepackage{stfloats}
\usepackage{url}
\usepackage{verbatim}
\usepackage{graphicx}
\usepackage{cite}
\begin{document}

\title{Improving Diffusion Models for ECG Imputation with an Augmented Template Prior}

\author{Alexander Jenkins$^{*, 1}$\thanks{$^*$ Equal contribution.}\thanks{$^1$ Department of Electrical and Electronic Engineering, Imperial College London, London, United Kingdom.}, Zehua Chen$^{*, 1}$, Fu Siong Ng$^2$\thanks{$^2$ National Heart and Lung Institute, Imperial College London, London, United Kingdom.}, Danilo Mandic$^{\dag, 1}$\thanks{$\dag$ Correspondence to: Danilo Mandic $<$d.mandic@imperial.ac.uk$>$.},~\IEEEmembership{Fellow,~IEEE}}
        % <-this % stops a space

\markboth{Journal of \LaTeX\ Class Files,~Vol.~14, No.~8, August~2021}%
{Shell \MakeLowercase{\textit{et al.}}: A Sample Article Using IEEEtran.cls for IEEE Journals}


\maketitle

\begin{abstract}
Pulsative signals such as the electrocardiogram (ECG) are extensively collected as part of routine clinical care. However, noisy and poor-quality recordings are a major issue for signals collected using mobile health systems, decreasing the signal quality, leading to missing values, and affecting automated downstream tasks. Recent studies have explored the imputation of missing values in ECG with probabilistic time-series models. Nevertheless, in comparison with the deterministic models, their performance is still limited, as the variations across subjects and heart-beat relationships are not explicitly considered in the training objective. In this work, to improve the imputation and forecasting accuracy for ECG with probabilistic models, we present a template-guided denoising diffusion probabilistic model (DDPM), PulseDiff, which is conditioned on an informative prior for a range of health conditions. Specifically, 1) we first extract a subject-level pulsative template from the observed values to use as an informative prior of the missing values, which personalises the prior; 2) we then add beat-level stochastic shift terms to augment the prior, which considers variations in the position and amplitude of the prior at each beat; 3) we finally design a confidence score to consider the health condition of the subject, which ensures our prior is provided safely. Experiments with the PTBXL dataset reveal that PulseDiff improves the performance of two strong DDPM baseline models, CSDI and SSSD$^{S4}$, verifying that our method guides the generation of DDPMs while managing the uncertainty. When combined with SSSD$^{S4}$, PulseDiff outperforms the leading deterministic model for short-interval missing data and is comparable for long-interval data loss.
\end{abstract}


\section{Introduction}
The advent of wearable technology in mobile health (mHealth) has enabled passive real-time monitoring of personal health through recordings of physiological signals. Among these mHealth systems, pulsative signals such as electrocardiogram (ECG) \cite{harryecg,PulseImpute}, photoplethysmography (PPG) \cite{harryppg} and blood pressure \cite{zehuabloodpressure}, are extensively collected since their close relationship with the health conditions of both cardiovascular and circulatory system. However, in comparison with clinical monitoring systems, a major challenge in mHealth systems is the missing and corrupted values, for example due to poor contact with the electrodes during the recording of the ECG. Such issues can disrupt these recordings and deteriorate the performance of health monitoring algorithms.

\begin{figure*}[ht]
    \centering
    \includegraphics[width=\textwidth]{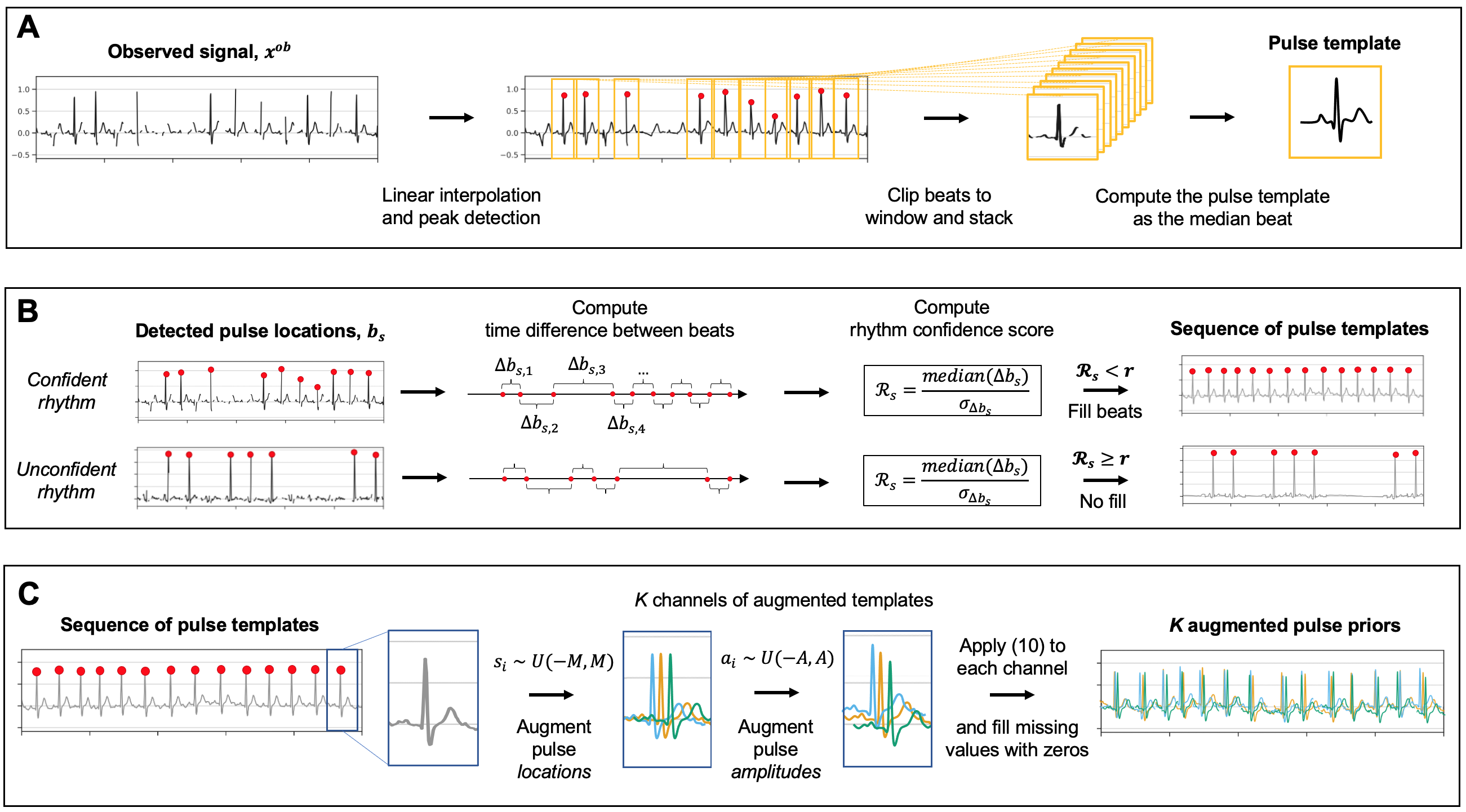}
    \caption{The proposed `PulseDiff' methodology to improve diffusion models for pulsative signal imputation. A) Calculation of the pulse template for each subject. B) Formation of the pulse prior using the rhythm confidence score to / to not fill missing beats. C) Augmentation of the pulse prior by applying random shifts in pulse location and amplitude.}
    \label{fig:proposed_method}
\end{figure*}

Recently, to promote the development of imputation methods for pulsative signals in mHealth systems, PulseImpute \cite{PulseImpute} has formulated a benchmark task where the ECG signals suffer two popular missing patterns in real life: transient (short-interval) missing values due to information packet loss; and extended (long-interval) missing values due to connectivity issues. PulseImpute provides a comprehensive study of existing deterministic methods, and establishes a strong baseline with an innovative transformer-based solution. Especially, it explicitly illustrates the difference between pulsative signals and other time-series data: $1$) pulsative signals have quasiperiodic structure with specific morphologies, e.g., the QRS complex in ECG; $2$) the specific morphologies vary over time and across populations, e.g., different morphologies shown by healthy subjects and patients. The first observation indicates that a pulsative signal is usually composed of different but similar repeating waveforms. Moreover, the second observation shows that the variances between these waveforms could be caused by diverse factors, such as subject age, gender, health condition, and the time of recording. However, during the imputation of pulsative signals, this additional information is usually unknown and the prediction only depends on the observed signals. As a result, deterministic methods for the imputation task may perform sub-optimally without this additional context information.

% The above analysis motivates our designs of pulsative signals imputation in this work: $1$) to improve the imputation accuracy, a probabilistic generative model may be useful to learn the randomness between beat-level samples of pulsative signals; $2$) a template of the beat samples in pulsative signal may be already a strong prior of the generation target, and could be taken as a condition to guide the generation process of a probabilistic model; $3$) the template in 

Recently, in comparison with deterministic prediction, probabilistic generative models, e.g., denoising diffusion probabilistic models \cite{DDPM,SGM,VariationalDiffusion} (DDPMs, diffusion models for short), have been popularly used in various tasks because of their superior performance in generation quality \cite{ResGrad,ImageDeblurring} and mode coverage \cite{SGM,DiffusionGAN}. Generally, diffusion models are composed of two processes and a model training stage: $1$) a forward process adds Gaussian noise on data samples to destroy them into the known prior distribution, i.e., Gaussian distribution; $2$) a reverse process gradually generates data from the prior distribution with iterative sampling process; $3$) an architecture-free network learns the gradient field (i.e., score function) of noisy samples in forward process, which guides the reverse process to reconstruct data samples. Given its probabilistic nature, it would be helpful to learn the randomness in pulsative signals, which is of importance to the imputation accuracy. Moreover, diffusion models have demonstrated strong results in time-series imputation and forecasting \cite{CSDI,SSSD}, while their advantages for pulsative signals have not been explored. Yet, a probabilistic generative model may be useful for the pulse imputation task as they could model the stochastic components of these signals, such as varying morphologies and a lack of context information.

%However, in existing methods and the solution proposed by PulseImpute, they expect to implicitly learn the quasi-periodic structure and the variations in morphologies from training dataset, and provide the imputation results in a deterministic way.

\begin{figure*}[ht]
    \centering
    \includegraphics[width=\textwidth]{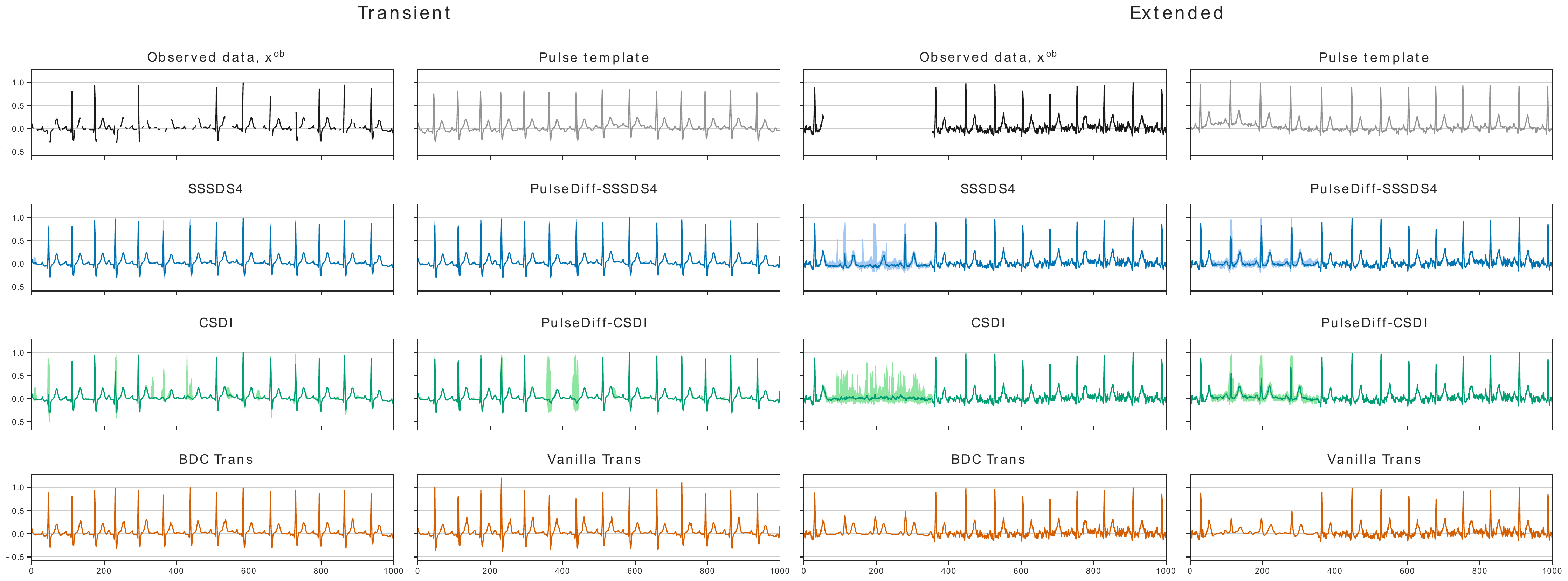}
    \caption{Comparison of imputation results on ECG waveforms from the \textsc{PTB-XL} dataset with 30\% transient and extended missing values. PulseDiff improves performance of conditioning denoising diffusion probabilistic models CSDI~\cite{CSDI} and SSSD$^{S4}$~\cite{SSSD} by leveraging a pulse template as additional conditioning information. Imputation performance with PulseDiff-SSSD$^{S4}$ also beats the state-of-the-art transformer model, BDC Trans, for transient missingness and is comparable for extended missingness.}
    \label{fig:imputation_results_30percent}
\end{figure*}

In this work, we introduce diffusion-based probabilistic generative models to the benchmark pulse imputation task. Moreover, to further boost the accuracy, we design PulseDiff, a diffusion model conditioned on an augmented template prior. Specifically, $1$) considering the quasiperiodic structure, we propose to extract a pulse template from the observed values, compute a template-based prior for missing values, and introduce a rhythm confidence score to reject misleading priors; $2$) considering the pulse morphologies varying over time, we augment our prior by adding random shifts to template positions and amplitudes; $3$) considering the pulse morphologies varying across populations, we extract the template at the subject-level to capture the subject-specific pulse morphology. Our methodology is highlighted in Figure \ref{fig:proposed_method}, and our improvements on generation results can be seen by a sample in Figure \ref{fig:imputation_results_30percent}.

 By designing the augmented template-based prior as the indicating information, we make the following contributions for pulse imputation:

\begin{itemize}
    \setlength{\itemsep}{0pt}
    \setlength{\parskip}{0pt}
    \item We introduce probabilistic generative models into the benchmark pulse imputation task. With conditional diffusion models, we achieve comparable results with the SOTA transformer-based baseline models. 
    \item We condition diffusion models on a data-driven prior to improve the imputation accuracy. For each individual, we formulate the methods to extract pulse template and compute the prior for missing values. Then, we design conditioning augmentation methods and confidence scores to strengthen the prior. 
    \item We demonstrate that our method steadily improves the reconstruction accuracy of two baseline diffusion models. Moreover, our model outperforms both deterministic and probabilistic works in pulse imputation of transient (short-interval) missing values, and strengthens the probabilistic models during imputation of extended (long-interval) missing values.    
\end{itemize}

\section{Related Works}
\subsection{Pulse Imputation}
The benchmark task of pulsative signal imputation is established by PulseImpute~\cite{PulseImpute}. They provide a comprehensive study which contains both signal processing methods such as mean filling~\cite{meanfilling}, linear interpolation~\cite{linearinterpo}, and Fast Fourier Transformer (FFT) imputation~\cite{FFT}, and deep learning methods such as $1$) RNN imputation benchmark BRITS~\cite{BRITS}, $2$) transformer-based architectures DeepMVI~\cite{DeepMVI}, Vanilla Transformer~\cite{Attention}, and Conv9~\cite{Conv9}, and $3$) GAN-based approaches BRITS w/ GAIL~\cite{GAIL,NAOMI} and NAOMI~\cite{NAOMI}. As these methods are not designed to capture the quasiperiodicity and morphologies of pulsative signals, their imputation results are limited. Therefore, PulseImpute~\cite{PulseImpute} specifically designs a bottleneck dilated convolutional transformer for pulse imputation, BDC Trans, and achieves the SOTA performance. However, probabilistic models are not considered in PulseImpute. In this work, we establish the probabilistic pulse imputation baselines, and propose improving techniques which are specifically designed for pulsative signals.

\subsection{Denoising Diffusion Probabilistic Models}
The probabilistic generative model DDPMs have achieved the state-of-the-art generation quality in both generation tasks, e.g., image generation~\cite{CascadedDDPM,DALLE2,Imagen}, video generation~\cite{MakeAVideo,ImagenVideo} and waveform generation~\cite{PriorGrad,InferGrad,BinauralGrad}, and conditional restoration tasks, e.g., image restoration~\cite{ImageDeblurring}, sketch-guided image generation~\cite{SDEdit}, and time-series imputation~\cite{CSDI}. Especially, in time-series modelling, diffusion models have shown strong capability in various datasets such as Traffic, Air Quality, and Healthcare. 

However, the advantages of diffusion models have not been widely explored for bio-electrical signal processing. DeScoD-ECG~\cite{DeScoD-ECG} employs diffusion models for ECG denoising. One related work is SSSD~\cite{SSSD}. SSSD utilized the generation quality of diffusion models~\cite{DiffWave,CSDI} and improved the network architecture with state-space models~\cite{SaShimi} in order to capture long-term dependencies. However, they focused on general time-series imputation and forecasting, and ECG imputation is simply considered as one of the application scenarios. In comparison, our PulseDiff is specifically designed for pulse imputation by considering the quasiperiodic structure and varying morphologies, and is free to the choice of foundation diffusion models. 

\section{Methodology}
\subsection{Preliminary}
\paragraph{Probabilistic Pulse Imputation}
The task of pulse imputation aims to generate the missing values $x^{ta}$, given the observed values $x^{ob}$. In traditional deterministic models, a mapping function $x^{ta}=f_{\theta}(x^{ob})$ is learned to estimate missing values with convolutional or transformer based architectures. In comparison, with probabilistic models, we are estimating the true conditional data distribution $p(x^{ta}_{0}|x^{ob})$ with the model distribution $p_{\theta}(x^{ta}_{0}|x^{ob})$.

To simplify annotation, we follow the notation introduced in the time-series diffusion model CSDI~\cite{CSDI}, to denote each pulse recording as $\{x,m,s\}$. The pulse values are stored in $x\in\mathbb{R}^{L}$, where $L$ is the time length of recording. The missing or observed pulse values are denoted by an observation mask $m\in\mathbb{R}^{L}$, where $m_{l}=0$ if $x_{l}$ is missing and $m_{l}=1$ if $x_{l}$ is observed. The timestamp of values is stored by $s\in\mathbb{R}^{L}$. Therefore, for each single-channel pulse recording, we denote it by $\{x,m,s\}$, and $x$ is composed by missing values $x^{ta}$ and observed values $x^{ob}$.

\vspace{1em}
\paragraph{Conditional Diffusion Models}
\label{sec:cond_diffusion_models}
For pulse imputation, we introduce conditional diffusion models which use a forward process to destroy the pulsative signals $x_{0}$ into standard Gaussian noise $x_{T}\sim\mathcal N(0,I)$ with a predefined noise schedule $ 0 < \beta_{1} < \dots < \beta_{t} < \dots <  \beta_{T} < 1$, and use a reverse process to iteratively generate the missing values $x^{ta}$ in $x_{0}$. In forward process, the transition probability and the noise distribution $q(x_{t}|x_{0})$ at each time step $t\in [1,\dots,T]$ can be denoted as:
\begin{align}
q(x_{t}|x_{t-1})&=\mathcal{N}(x_{t};\sqrt{1-\beta_{t}}x_{t-1},\beta_{t}I), \\
q(x_{t}|x_{0})&=\mathcal  N(x_{t};\sqrt{\bar{\alpha}_{t}}x_{0},(1-\bar{\alpha}_{t})\epsilon),
\end{align} 

where $\epsilon\sim\mathcal N(0,I)$ is injected standard Gaussian noise, $\alpha_{t}:=1-\beta_{t}$ and $\bar{\alpha}_{t}:=\prod_{s=1}^{t}\alpha_{s}$ represent the noise level at time step $t$. In reverse process, a denoising process starting from $p(x_{T})\sim\mathcal N(0,I)$ gradually removes the noise at each inference step given observed data $x^{ob}$: 
\begin{align}
p_{\theta}(x_{0:T})&=p(x_{T})\prod_{t=1}^{T}p_{\theta}(x_{t-1}|x_{t},x^{ob}), \\
p_{\theta}(x_{t-1}|x_{t})&=\mathcal{N}(x_{t-1};\mu_{\theta}(x_{t},t,x^{ob}),\sigma_{\theta}^{2}I),
\end{align}
where the mean and the variance can be parameterized as:
\begin{align}
\mu_{\theta}(x_{t},t,x^{ob})&=\frac{1}{\sqrt{\alpha_{t}}}(x_{t}-\frac{\beta_{t}}{\sqrt{1-\bar{\alpha}_{t}}}\epsilon_{\theta}(x_{t},t,x^{ob})),
\\
\sigma_{\theta}^{2}&=\frac{1-\bar{\alpha}_{t-1}}{1-\bar{\alpha}_{t}}\beta_{t}.
\label{sigma}
\end{align}
Diffusion models are trained by maximizing the variation lower bound of the likelihood $p_{\theta}(x_{0})$. With the same parameterization of $\mu_{\theta}$, we employ the training objective from \cite{DDPM,DiffWave,CSDI}:
\begin{align}
\label{trainingobjective}
L(\theta)=\mathbb{E}_{x_{0},\epsilon,t}\left \| \epsilon - \epsilon_{\theta}(\sqrt{\bar{\alpha}_{t}}x_{0}+\sqrt{1-\bar{\alpha}_{t}}\epsilon,t,x^{ob}) \right\|^2_{2}.
\end{align}

\subsection{PulseDiff}
The proposed work utilizes a data-driven template to instruct the generation process of diffusion models. To improve the imputation accuracy, our design deals with several considerations: $1$) how to construct the template? For example, either a population average (global) template or an individual template may be helpful; $2$) how to consider the unknown health condition of each recording? Given same $x^{ob}$, the ground-truth values can be distinctively different due to the patients health condition; $3$) how to instruct diffusion models using the template? For example, a small shift between the template and the ground-truth data may induce error in imputation results.

In this section, the methods to compute a pulse prior, $p$, from an individual recording $\{x,m,s\}$ are first introduced. To calculate an effective pulse prior, we follow a three-step process that accounts for the quasi-periodic structure of pulsative signals, subject-specific pulse morphologies, and the temporal variability in pulse occurrence. This process is illustrated in figure \ref{fig:proposed_method}.

% \subsubsection{template Extraction}
% \label{sec:TC}
\vspace{1em}
\paragraph{Pulse Template}
In order to compute a pulse prior, $p\in \mathbb{R}^L$, for each individual recording, $\{x,m,s\}$, a pulse template will first be extracted from the observed values, $x^{ob}$. Given that pulsative morphologies vary across the population, a pulse template will be extracted for each subject, such that the individual's morphology is captured. The quasi-periodic structure of pulsative signals serves as the foundation for extracting a pulse template waveform from them, which can be seen as capturing the common information of each beat. To create a pulse template, pulse waveforms will be detected from the partially observed recording and averaged. To enable the detection of pulse waveforms in the recording using signal processing algorithms, the observed values, $x^{ob}$, will be linearly interpolated to form $x^{ob-l}$; a new sequence with no missing values. 

To detect pulse waveforms in $x^{ob-l}$, a matched filter will be used. As the interpolated waveform in $x^{ob-l}$ is likely to be missing important waveform information, such as peaks, a matched filter is chosen due to its robustness for identifying patterns in signals. For use as the matched filter template, an external pulse template, $p_{ext}$, is constructed from a small group of recordings, $N_{ext}$, split from the training dataset of size $N$, whereby $N_{ext} \ll N$. To form the external pulse template for ECG recordings, a Christov detector \cite{Christov2004} will be used to detect the locations of each QRS complex present in the $N_{ext}$ fully-observed recordings. A subtraction of $250$ ms is applied to the detected locations to include the P-wave. Each signal is split based on the detected locations, and subsequently overlayed, to compute the external pulse template as the median average waveform.

The cross correlation is given by,
\begin{equation}
\label{eqn:mf-ext}
c(t) = x(t) \ast h(-t) = \sum_m x(t+m)h(m),
\end{equation}
where $\ast$ represents the convolution of a signal $x(t)$ and a time-reversed matched filter template $h(-t)$. The cross-correlation coefficient, $c_{ext}(t)$, between the signal $x^{ob-l}(t)$ and the external pulse template, $p_g(t)$, is computed from (\ref{eqn:mf-ext}). The value $c_{ext}(t)$ will then be thresholded to detect pulse waveforms in $x^{ob-l}(t)$. Specifically, values of $c_{ext}(t)$ above the $97^{th}$ percentile and separated by at least $400$ ms (assuming a maximum detectable resting heart rate of $150$ bpm) are classified as the locations of pulse waveforms. The thresholds were decided from discussions with a clinical expert. As a result, a sequence of detected beat locations is formed. An intermediate pulse template, $p_{s}^\prime$, will then be constructed by splitting the observed signal, $x^{ob}$, at the detected locations, overlaying these segments, and computing the median average waveform.

To further improve the accuracy of beat detection, and thus the pulse template, the process detailed in the previous paragraph will be repeated using $p_{s}^\prime(t)$ as the matched filter template. The result is a sequence of detected pulse locations, $b_{s} = (b_{s,n})_{n \in \mathbb{N}, 0 \leq b_{s, n} < L}$, where $b_{s,n}$ is the index of the $n^{th}$ beat. By splitting $x^{ob}$ at $b_{s}$ and overlaying these segments, the final pulse template, $p_{s}$, is computed as the median average waveform. By using this two-step detection method, subject-specific pulse morphologies have been leveraged to detect beats and create the final pulse template.

In summary, for each recording, a pulse template, $p_{s}$, is extracted using the following process:
\begin{enumerate}
    \item The observed values, $x^{ob}$, will be linearly interpolated to form $x^{ob-l}$, a new sequence with missing values filled.
    \item A sequence of pulse locations, $b_{s} = (b_{s,n})_{n \in \mathbb{N}, 0 \leq b_{s, n} < L}$, will be detected in $x^{ob-l}$ using the Christov detector \cite{Christov2004} and a matched filter.
    \item The interpolated observations, $x^{ob-l}$, are split into windows around each detected pulse. 
    \item The windowed signals are stacked and the median pulse waveform is computed. This will be used as the individual's pulse template, $p_{s}$.
\end{enumerate}

\vspace{1em}

\paragraph{Rhythm Confidence Score}
An individual could have a health condition such as a cardiac arrhythmia which causes an irregular pulse rhythm. However, information regarding the rhythm could be used to strengthen the template prior. For example, if a recording was known to have a regular rhythm, it would be possible to extend the detected pulse locations in $b_s$ by predicting beat locations at the same rhythm.

To exploit the quasi-periodic structure of pulsative signals, we introduce a score to quantify our confidence in the observed rhythm that depends on each recording $\{x,m,t\}$. The rhythm confidence score will be used to accurately impute missing beat locations, when the health condition is unknown during imputation.

The rhythm confidence score is derived from the ordered sequence of pulse locations, $b_{s} = (b_{s, n})_{n \in \mathbb{N}, 0 \leq b_{s,n} < L}$. Specifically, a new sequence, $\Delta b_{s} = (\Delta b_{s, n})_{n \in \mathbb{N}}$, is constructed where each element $\Delta b_{s, n}$ is the inter-beat interval such that $\Delta b_{s, n} = b_{s, n+1} - b_{s, n}$. From this, the median inter-beat interval, $median(\Delta b_s)$, and the standard deviation of inter-beat intervals, $\sigma_{\Delta b_s}$, will be computed. The rhythm confidence score for a subject is given by,
\begin{equation}
    \label{eqn:rhythm_conf}
    \mathcal{R}_s = \frac{\sigma_{\Delta b_s}}{median(\Delta b_s)}.
\end{equation}
This has a form similar to the coefficient of variation, $CV=\frac{\sigma}{\mu}$, but median is used in place of the mean $\mu$ to provide a more robust estimate of the average beat interval in the presence of missing values. 

The rhythm confidence score will be thresholded to specify if missing pulse locations should be imputed with a pulse template, which could strengthen the pulse prior. Missing beat locations will be imputed using the median inter-beat interval as detailed in Section 2 of the supplementary materials. The total sequence of pulse locations including imputed beats will be referred to as $b_{s}^\prime = (b_{s, n})_{n \in \mathbb{N}, 0 \leq b_{s,n}^\prime < L}$, where $b_{s,n}^\prime$ is the index of the $n^{th}$ beat, and the length of $b_{s}^\prime$, $n^\prime$ exceeds $b_{s}$, $n$, i.e. $n^\prime > n$. Using the rhythm confidence score, pulse templates, $p_{s}$, are placed in sequence at all detected beat locations to form a pulse prior as
\begin{equation}
    p(t) =
    \begin{cases}
    \sum_{i=1}^{n^{\prime}} p_s(t - b_{s,i}^\prime), & \text{if } \mathcal{R}_s < r \\
    \sum_{i=1}^n p_s(t - b_{s,i}), & \text{otherwise,}
    \end{cases}
\end{equation}
whereby a value of $r=1$ will be fixed during this work, implying the variability of observed intervals must be less than the median interval for pulse templates to be placed at missing pulse locations.

Due to the fact $p_s$ is of length one beat, and smaller than the length $L$ of the recording, the sequence of pulse templates may be non-overlapping. Therefore, a complete pulse prior is formed by filling the missing values with zeros. This interpolated waveform will here-in be referred to as the `fixed pulse prior'.

\vspace{1em}
\paragraph{Template Augmentation}
If the beat locations in $b_{s}$ and $b_{s}^\prime$ vary from the ground-truth then the template could provide misleading information. The amount of variation between predicted and ground truth beat locations is driven by the following factors: 1) uncertainty in beat detection, 2) uncertainty in beat imputation, 3) uncertainty in pulse morphology (e.g. amplitude), and 4) uncertainty due to natural heart rate variability. To model these uncertainties, we propose to augment the pulse prior by adding random shifts to the beat location and amplitude.

An augmented pulse prior will be formed as
% \begin{equation}
% p(t) = \sum_{i=1}^n (a_i + p_s(t - b_i)),
% \end{equation}
\begin{equation}
    p(t) =
    \begin{cases}
    \sum_{i=1}^{n^{\prime}} (a_i + p_s(t - b_{s,i}^\prime - s_i)), & \text{if } \mathcal{R}_s < r \\
    \sum_{i=1}^n (a_i + p_s(t - b_{s,i} - s_i)), & \text{otherwise,}
    \end{cases}
    \label{eqn:augmented_pulse_prior}
\end{equation}
where $p_s(\cdot)$ represents the pulse template positioned at a randomly shifted beat location. Specifically, the random shift in beat location, $s_i$, is modelled as an i.i.d random variable drawn from a uniform distribution, i.e., $s_i \sim U(-M, +M)$, and is used to augment the location of the beat. Furthermore, each $a_i$ represents a random shift in amplitude for each beat, modelled as an i.i.d random variable drawn from a uniform distribution, i.e., $a_i \sim U(-A, A)$. Again, missing values are filled with zeros to avoid providing potentially misleading values between the beats.

In order to provide better coverage of the ground truth data, we propose to sample $K$ augmented pulse priors for use as conditioning information, with each augmented prior denoted as $p_k$ for $k = 1,2...,K$ and calculated using (\ref{eqn:augmented_pulse_prior}). The result is an augmented template prior, $\mathbf{p}\in\mathbb{R}^{K\times L}$, for each individual, with $p_k$ stored in each row.

\vspace{1em}
\paragraph{Template-Conditional Diffusion Models}
The data for an individual can now be represented as $\{x, m, s, \mathbf{p}\}$, where $\mathbf{p}\in\mathbb{R}^{K\times L}$ is their augmented template prior. The augmented template prior is used as additional conditioning information such that the reverse process is reformulated as,

\begin{align}
p_{\theta}(x^{ta}_{0:T}|x^{ob},\mathbf{p})&=p(x^{ta}_{T})\prod_{t=1}^{T}p_{\theta}(x^{ta}_{t-1}|x^{ta}_{t},x^{ob},\mathbf{p}), \label{eqn:tdiff_reverse1}\\
p_{\theta}(x^{ta}_{t-1}|x^{ta}_{t},x^{ob},\mathbf{p})&=\mathcal{N}(x^{ta}_{t-1};\mu_{\theta}(x^{ta}_{t},t|x^{ob},\mathbf{p}),\sigma_{\theta}^{2}I),
\label{eqn:tdiff_reverse2}
\end{align}
where $x^{ta}_{T}\sim\mathcal N(0,I)$. The variance $\sigma_{\theta}^{2}$ is defined the same as in (\ref{sigma}), while the mean function is parameterized as, 
\begin{align}
\mu_{\theta}(x^{ta}_{t},t|x^{ob},\mathbf{p})=\frac{1}{\sqrt{\alpha_{t}}}(x^{ta}_{t}-\frac{\beta_{t}}{\sqrt{1-\bar{\alpha}_{t}}}\epsilon_{\theta}(x^{ta}_{t},t|x^{ob},\mathbf{p})).
\end{align}
The modified training objective is given by,
\begin{align}
\label{trainingobjective}
L(\theta)=\mathbb{E}_{x_{0},\epsilon,t}\left \| \epsilon - \epsilon_{\theta}(\sqrt{\bar{\alpha}_{t}}x_{0}+\sqrt{1-\bar{\alpha}_{t}}\epsilon,t,x^{ob},\mathbf{p}) \right\|^2_{2}.
\end{align}
Given the function $\epsilon_{\theta}$, observed data $x^{ob}$ and the augmented template prior $\mathbf{p}$, we can sample $x_0^{ta}$ using the reverse process in (\ref{eqn:tdiff_reverse1}) and (\ref{eqn:tdiff_reverse2}) for pulse imputation with PulseDiff.

\section{Experiments}
\label{Experiments}

\subsection{Experiment Setup} 
\paragraph{Training Dataset}
We use the open-source PTB-XL dataset\footnote{\url{https://physionet.org/content/ptb-xl/1.0.3/}} for ECG imputation. PTB-XL contains both the recordings from the collection of with co-occurring pathologies, e.g., myocardial infarction and conduction disturbance, and the samples from healthy populations. The time length of PTB-XL recordings is $10$ seconds. The sampling rate is $100$Hz. To align with PulseImpute, we follow their methods for splitting the data: $8730$ samples for training, $2176$ samples for validation, and $10931$ samples for testing. In our experiments, we randomly select $10$ recordings from the training set to calculate the global template $p_{g}$ and use the remaining $8720$ samples for training. We use $64$ samples for validation in the training process of diffusion-based models, and randomly extract $100$ samples as the evaluation set for both PulseImpute baseline models and diffusion-based models.  

\vspace{1em}
\paragraph{Training Details}
\label{sec:training_details}
For transformer-based models built by PulseImpute~\cite{PulseImpute}, we utilize their open-source implementation\footnote{\url{https://github.com/rehg-lab/pulseimpute}} and pretrained models. Training and data details can be found in their paper and supplementary materials. For diffusion-based probabilistic generative models, we establish two baseline models with CSDI\footnote{\url{https://github.com/ermongroup/CSDI}} and SSSD$^{S_4}$ \footnote{\url{https://github.com/AI4HealthUOL/SSSD}} by incorporating their official implementation into pulse imputation. We denote the template-guided ones as PulseDiff-CSDI and PulseDiff-SSSD$^{S_4}$, respectively.

For CSDI and PulseDiff-CSDI, we employ the same settings as used in CSDI~\cite{CSDI}. Specifically, the number of residual layers is set as $4$, the number of residual channels as $64$, the dimension of the diffusion embedding as $128$, and the number of time steps in both diffusion and reverse process as $T=50$ with a quadratic noise schedule from $\beta_{1}=0.0001$ to $\beta_{T}=0.5$. A total of $200$ epochs are used and batch size of $16$. As we focus on the single-channel pulse recording, we remove the feature transformer layer and retain the temporal transformer layer with embedding dimension as $16$ and $8$ self-attention heads. The Adam optimizer is employed. The learning rate is $1e^{-3}$ and decayed to $1e^{-4}$ at $75\%$ and $1e^{-5}$ at $90\%$ of total epochs.

For SSSD$^{S_4}$ and PulseDiff-SSSD$^{S_4}$, we employ the same settings as used in SSSD$^{S_4}$~\cite{SSSD}. Specifically, the number of residual layers is set as $36$, the number of residual channels as $256$, number of skip channels as $256$, diffusion embedding dimension $1$ as $128$, embedding dimensions $2$ and $3$ as $256$, and the number of time steps in both diffusion and reverse process as $T=200$ with a linear noise schedule from $\beta_{1}=0.0001$ to $\beta_{T}=0.02$. A total of $200$ epochs are used and batch size of $4$. The Adam optimizer is employed and the learning rate is $2e^{-4}$ with no decay. We train each probabilistic generative model on a single Nvidia Tesla A30 GPU.

\begin{table*}[ht]
\scriptsize 
\centering
\caption{Imputation performance on \textsc{ptbxl} dataset and downstream tasks with 30\% \textbf{transient} missing value ratio. MSE is used for reconstruction accuracy, and other metrics are used in downstream tasks: classification and beat detection.}
\begin{tabular}{l|c|cccccc}
\toprule
Model &   MSE $\downarrow$ &  Rhythm AUC $\uparrow$ &  Form AUC $\uparrow$ &  Diagnostic AUC $\uparrow$ &    F1 $\uparrow$ &  Precision $\uparrow$ &  Sensitivity $\uparrow$ \\
\midrule
               BRITS w/ GAIL \cite{GAIL} & 0.0346 &      0.8673 &    0.7814 &          0.7981 & 0.4638 &     0.6588 &       0.3579 \\
               \textbf{Template} & 0.0215 &      0.9367 &    0.7425 &          0.8418 & 0.6852 &     0.7471 &       0.6328 \\
               NAOMI \cite{NAOMI} & 0.0125 &      0.9388 &    0.8411 &          0.8528 & 0.7709 &     0.7612 &       0.7809 \\
              Vanilla Trans \cite{Attention} & 0.0081 &      0.9372 &    0.8205 &          0.8674 & 0.8380 &     0.8176 &       0.8594 \\
               Conv9 Trans \cite{Conv9} & 0.0081 &      0.9236 &    0.8305 &          0.8731 & 0.8165 &     0.8039 &       0.8295 \\
               BDC Trans \cite{PulseImpute} & 0.0043 &      0.9429 &    0.8129 &          0.8745 & 0.8995 &     0.8932 &       0.9058 \\
\midrule
               CSDI \cite{CSDI} & 0.0091 &      0.9356 &    0.8280 &          0.8610 & 0.7631 &     0.8388 &       0.7000 \\
               \textbf{PulseDiff-CSDI} & 0.0065 &      0.9511 &    0.8309 &          0.8740 & 0.8507 &     0.8780 &       0.8250 \\
\midrule
               SSSD$^{S4}$ \cite{SSSD} & 0.0023 &      0.9509 &    \textbf{0.8450} &          0.8734 & 0.9304 &     0.9457 &       0.9155 \\
               \textbf{PulseDiff-SSSD$^{S4}$} & \textbf{0.0021} &      \textbf{0.9546} &    0.8377 &          \textbf{0.8822} & \textbf{0.9402} &     \textbf{0.9626} &      \textbf{ 0.9187 }\\
\bottomrule
\end{tabular}
\label{tab:results-transient-ptbxl}
\end{table*}

\begin{table*}[ht]
\scriptsize 
\centering
\caption{Imputation performance on \textsc{ptbxl} dataset and downstream tasks with 30\% \textbf{extended} missing value ratio. MSE is used for reconstruction accuracy, and other metrics are used in downstream tasks: classification and beat detection.}
\begin{tabular}{l|c|cccccc}
\toprule
Model &   MSE $\downarrow$ &  Rhythm AUC $\uparrow$ &  Form AUC $\uparrow$ &  Diagnostic AUC $\uparrow$ &    F1 $\uparrow$ &  Precision $\uparrow$ &  Sensitivity $\uparrow$ \\
\midrule
               BRITS w/ GAIL \cite{GAIL} & 0.0536 &      0.8868 &    0.7714 &          0.8007 & 0.0767 &     0.0758 &       0.0775 \\
               NAOMI \cite{NAOMI} & 0.0400 &      0.9104 &    0.7737 &          0.8403 & 0.1593 &     0.1653 &       0.1538 \\
               \textbf{Template} & 0.0363 &      0.9276 &    \textbf{0.8372} &          0.8547 & 0.4704 &     0.4766 &       0.4644 \\
               Conv9 Trans \cite{Conv9} & 0.0307 &      0.9114 &    0.7731 &          0.8143 & 0.0243 &     \textbf{0.9000} &   0.0123 \\
               Vanilla Trans \cite{Attention} & 0.0248 &      0.8980 &    0.7736 &          0.7967 & 0.4581 &     0.6156 &       0.3648 \\
               BDC Trans \cite{PulseImpute} & \textbf{0.0172} &      \textbf{0.9597} &    0.7908 &          0.8074 & \textbf{0.6582} &     0.7155 &       \textbf{0.6094} \\
\midrule
               CSDI \cite{CSDI} & 0.0306 &      0.8835 &    0.7673 &          0.8214 & 0.0328 &     0.2583 &       0.0175 \\
               \textbf{PulseDiff-CSDI} & 0.0236 &  0.9272 &    0.8105 &   0.8395 & 0.4961 & 0.6030  &   0.4213 \\
\midrule
               SSSD$^{S4}$ \cite{SSSD} & 0.0197 &     0.9331 &    0.8038 &          \textbf{0.8640} & 0.6008 &     0.7216 &       0.5147 \\
               \textbf{PulseDiff-SSSD$^{S4}$} & 0.0192 &      0.9316 &    0.8030 &          0.8528 & 0.6083 &    0.7602 &       0.5070 \\
               
\bottomrule
\end{tabular}
\label{tab:results-extended-ptbxl}
\end{table*}

PulseDiff-CSDI and PulseDiff-SSSD$^{S_4}$ introduce three additional hyperparameters for tuning the augmented pulse templates: the magnitude of random shifts in beat location, $M$; the magnitude of random shifts in beat amplitude, $A$; and the total number of augmented pulse priors, $K$. Hyperparameter search is conducted using a Bayesian optimisation with 50 iterations, each for a total of $25$ epochs, whereby continuous values of $M$ and $A$ are trialled in the interval $[0, 10]$ and integer values of $K$ in the interval $[1, 16]$. The optimal hyperparameter values are chosen to minimize the MSE between imputed values and ground-truth on the validation set. Because PulseDiff is free to the choice of foundation diffusion model, we also trial providing the augmented template prior as either input only, conditioning side information only, or as both; for compatibility with the DiffWave~\cite{DiffWave} architectures of CSDI and SSSD$^{S_4}$.

\vspace{1em}
\paragraph{Missingness Patterns}
Following PulseImpute, we use two types of missingness patterns corresponding to short and long intervals of data loss: \textit{transient} missingness which models the sporadic loss of $50$ ms information packets, and \textit{extended} missingness which models sensor attachment issues. Transient and extended missingness are both parameterized by a missingness percentage that controls the proportion of removed samples in a waveform. A fixed missingness percentage of 30\% (most common missing percentage observed in mHealth systems) is used to train all models.

\vspace{1em}
\paragraph{Evaluation Methods}
Following PulseImpute, we evaluate the model performance using reconstruction accuracy and downstream tasks. For evaluation of reconstruction accuracy, we use the Mean Square Error (MSE) for both deterministic models and probabilistic generative models. Following imputation of missing values with each model, the reconstructed waveform was subject to downstream tasks. First, a multi-label classification task was conducted whereby diagnosis (e.g. WPW syndrome), form (e.g. inverted T-waves) and rhythm (e.g. atrial fibrillation) were predicted using three pre-trained xResNet1d's which were built by PulseImpute. Performance on the classification task is evaluated using macro-AUC. Finally, a beat detection task is performed on the imputed ECG waveforms. Specifically, beats were detected using the Stationary Wavelet Transform peak detector. The detected beat locations were compared to the ground truth beat locations, and beats in the imputed signal are matched to the true beats with a $50$ ms tolerance window. The beat detection measures of F1 score, precision, and recall, are used to evaluate performance on this task. Model performance is assessed for different missingness percentages, from 10\% to 50\% at a step size of 10\%, to evaluate the effectiveness of imputation methods in generalizing to varying amounts of missingness at evaluation stage. For comparison, the performance of PulseDiff will be evaluated against the (non-augmented) pulse prior, where imputed data is set equal to the prior, to demonstrate how PulseDiff leverages the prior to improve performance.

\vspace{1em}
\paragraph{Ablation Study}
To demonstrate the utility of each proposed component in PulseDiff, we will perform an ablation study. PulseDiff-CSDI will be gradually reduced to the baseline CSDI model through the following series of ablations: 1) remove the rhythm confidence score, 2) reduce the number of augmented templates from $K$ to $1$, 3) remove beat-level stochastic shift terms used in augmentation, 4) remove the pulse prior. By the final ablation, the model is reduced to CSDI. Each of these models will be setup as described in Section \ref{sec:training_details} and subject to the same downstream tasks. For computational ease, we evaluate these models with 50 training epochs on the data with 30\% \textit{transient missingness}. The evaluation dataset for the ablation study will be selected using stratified sampling with respect to the different class labels available. This is done because the PTB-XL dataset has unbalanced classes, but we aim to demonstrate the utility of PulseDiff across a complete range of patient conditions.

\begin{figure*}[h]
    \centering
    \includegraphics[width=\textwidth]{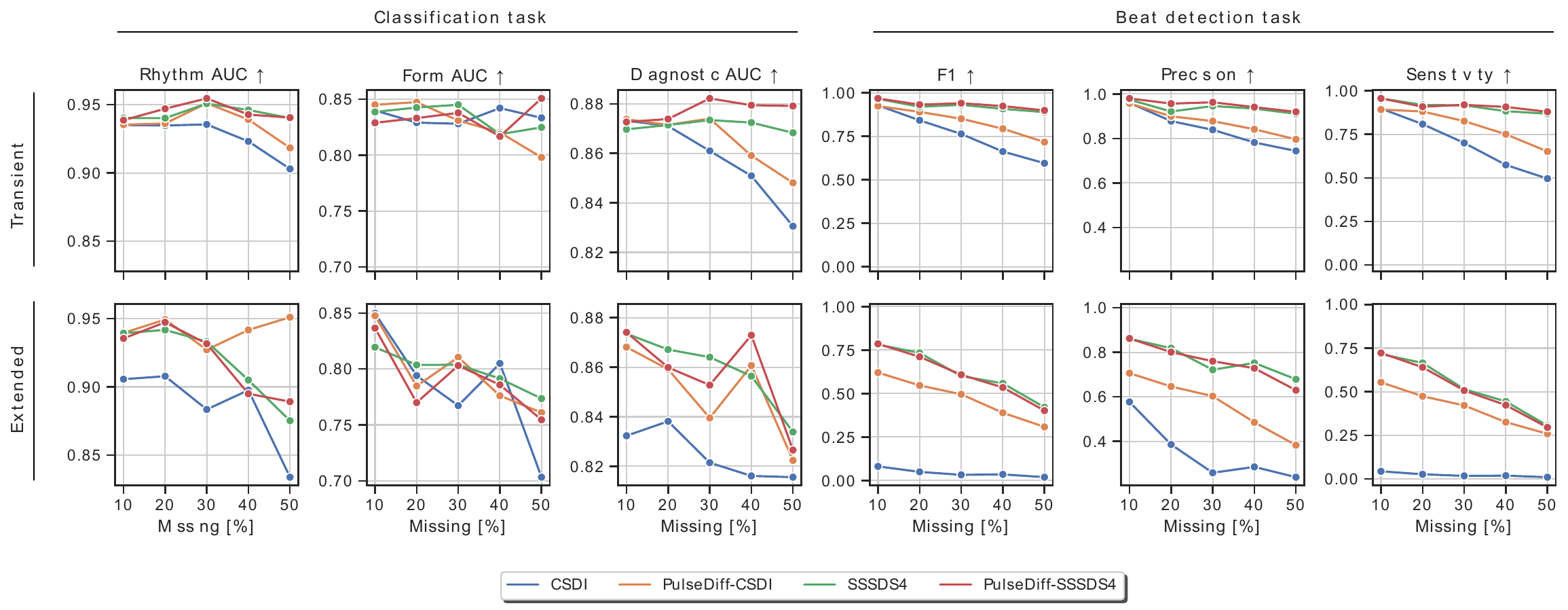}
    \caption{Performance on the classification and beat detection downstream tasks for probabilistic diffusion generative models for different percentages of transient and extended missingness.}
    \label{fig:classification_performance}
\end{figure*}

\subsection{Results}
\paragraph{Hyperparameters}
Bayesian optimisation on the validation set identified suitable hyperparameters for PulseDiff-CSDI to be location shift $M=2$, amplitude shift $A=0.01$ and number of templates $K=16$ for transient missingness, and $M=2$, $A=0.01$ and $K=16$ for extended missingness (where $M$ is rounded to nearest integer, and $A$ is shown for 2 decimal places). Providing the augmented template prior as both input and conditioning side information worked best for PulseDiff-CSDI. Similarly, the hyperparameters for PulseDiff-SSSD$^{S4}$ were found to be $M=1$, $A=0.00$ and $K=8$ for transient missingness, and $M=10$, $A=0.00$ and $K=8$ for extended missingness. Providing the augmented template prior as input only worked best for PulseDiff-SSSD$^{S4}$. PulseDiff-CSDI and PulseDiff-SSSD$^{S4}$ are trained as described in Section \ref{sec:training_details} using the identified hyperparameter setups, and their imputation performance are shown next. 

\begin{figure}[htp]
    \centering
    \includegraphics[width=\columnwidth]{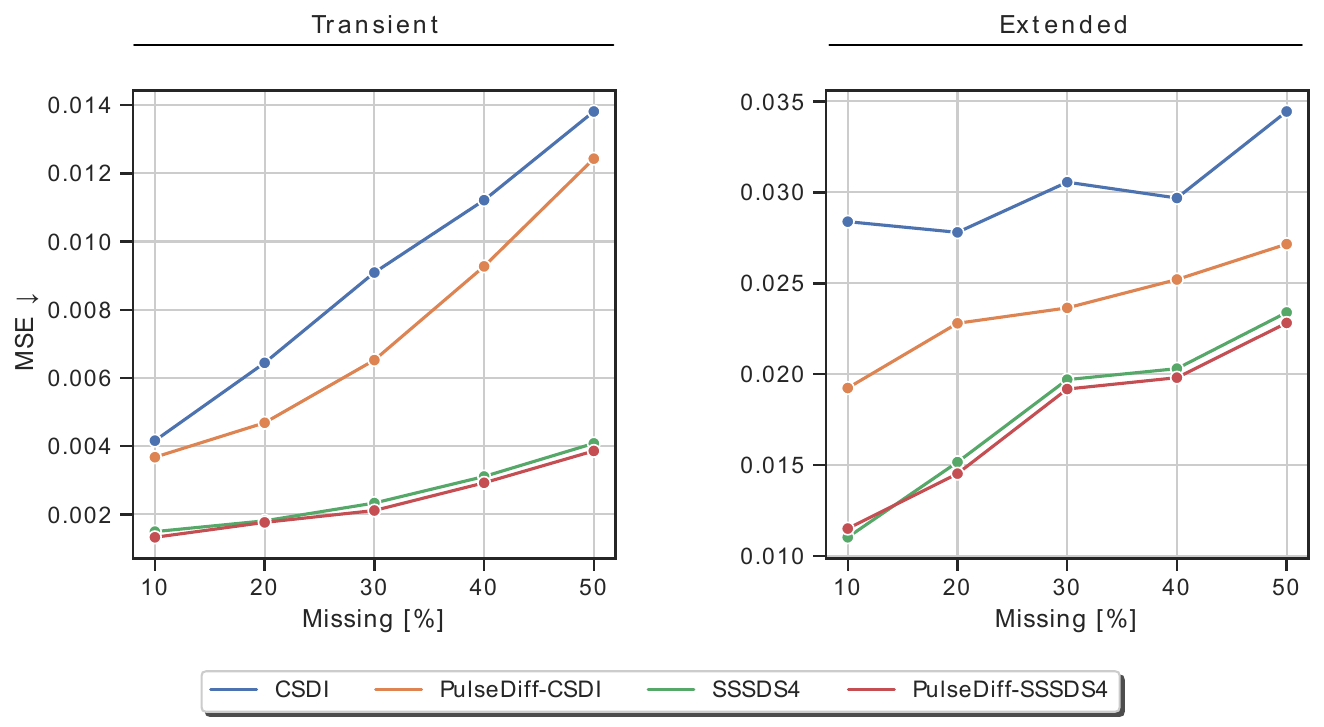}
    \caption{Reconstruction accuracy of probabilistic diffusion generative models as quantified using MSE for different percentages of transient and extended missingness.}
    \label{fig:mse_performance}
\end{figure}

\vspace{1em}
\paragraph{Reconstruction accuracy}
Accuracy of reconstruction is measured by the widely used MSE, the results of which for 30\% transient and extended missingness are quantified and displayed in Table \ref{tab:results-transient-ptbxl} and Table \ref{tab:results-extended-ptbxl}, respectively. From Table \ref{tab:results-transient-ptbxl} showing transient missingness, we can observe that our proposed method reduces the MSE for both CSDI and SSSD$^{S4}$. Moreover, \textit{PulseDiff-SSSD$^{S4}$ achieves SOTA result with an MSE of $0.0021$}, better-performing than all probabilistic and deterministic models. From Table \ref{tab:results-extended-ptbxl} showing extended missingness, we observe that our method also reduces the MSE for two diffusion baseline models. Moreover, PulseDiff-SSSD$^{S4}$ achieves the best accuracy among probabilistic models with an MSE of $0.0192$, which has been comparable to the SOTA quality achieved by BDC \cite{PulseImpute}. In addition, we notice that in comparison with our extracted pulse prior (Template in Table \ref{tab:results-transient-ptbxl} and Table \ref{tab:results-extended-ptbxl}), PulseDiff variants perform significantly better, verifying the advantage and necessity of probabilistic generative models. Given different missingness percentages, the reconstruction accuracy is also assessed for an overall evaluation of our methods for diffusion models. As shown in Figure \ref{fig:mse_performance}, our PulseDiff variants steadily decrease the MSE across all percentages (from 10\% to 50\%) of missing values tested.

% \begin{table*}[ht]
% \scriptsize 
% \centering
% \caption{Ablation study for PulseDiff components evaluated on the \textsc{ptbxl} dataset and downstream tasks with 30\% transient missing values. \\}
% \begin{tabular}{l|c|cccccc}
% \toprule
% Model &   MSE $\downarrow$ &  Rhythm AUC $\uparrow$ &  Form AUC $\uparrow$ &  Diagnostic AUC $\uparrow$ &    F1 $\uparrow$ &  Precision $\uparrow$ &  Sensitivity $\uparrow$ \\
% \midrule
% CSDI & 0.01490 &      0.90740 &    0.81950 &          0.85540 & 0.63210 &     \textbf{0.85210} &       0.50240 \\
% CSDI + 1 fixed prior & 0.01037 &     0.94016 &   0.81531 &         \textbf{0.88010} & 0.78004 &    0.80908 &      0.75302 \\
% CSDI + 1 fixed prior + score & 0.01019 &     0.94005 &   0.81489 &         0.87130 & 0.79257 &    0.83268 &      0.75616 \\
% CSDI + 16 augmented prior & 0.01138 &     \textbf{0.94113} &   \textbf{0.82922} &         0.87278 & 0.77533 &    0.82365 &      0.73237 \\
% \textbf{PulseDiff CSDI} = CSDI + 16 augmented prior + score & \textbf{0.01018} &     0.94034 &   0.82355 &         0.87422 & \textbf{0.80158} &    0.83589 &      \textbf{0.76997} \\
% \bottomrule
% \end{tabular}
% \label{tab:ablation_results}
% \end{table*}

\begin{table*}[ht]
\scriptsize 
\centering
\caption{Ablation study for PulseDiff components evaluated on the \textsc{ptbxl} dataset and downstream tasks with 30\% transient missing values. \\}
\begin{tabular}{l|c|cccccc}
\toprule
Model &   MSE $\downarrow$ &  Rhythm AUC $\uparrow$ &  Form AUC $\uparrow$ &  Diagnostic AUC $\uparrow$ &    F1 $\uparrow$ &  Precision $\uparrow$ &  Sensitivity $\uparrow$ \\
\midrule
CSDI \cite{CSDI} & 0.0177 &         0.8859 &       0.7344 &             0.7117 & 0.0245 &        0.0295 &          0.0209 \\
CSDI + 1 fixed prior & 0.0143 &      0.9254 &    0.7410 &          0.7208 & 0.0527 &     0.0462 &       0.0613 \\
CSDI + 1 fixed prior + score & 0.0122 &      0.9093 &    0.7532 &          0.7260 & 0.0425 &     0.0344 &       0.0556 \\
 CSDI + 16 augmented prior & 0.0120 &      0.9084 &    0.7619 &          0.7294 & 0.0499 &     0.0420 &       0.0613 \\
\textbf{PulseDiff-CSDI} = CSDI + 16 augmented prior + score & 0.0135 &      0.9079 &    0.7518 &          0.7250 & 0.0374 &     0.0320 &       0.0448 \\
\bottomrule
\end{tabular}
\label{tab:ablation_results}
\end{table*}

Figure \ref{fig:imputation_results_30percent} shows a case study of the reconstructed waveforms for 30\% transient and extended missingness for two different subjects. By leveraging a pulse prior, $p$, PulseDiff variants improve the imputation performance of CSDI and SSSD$^{S4}$ by a considerable margin. Upon comparison of with CSDI, PulseDiff-CSDI has improved the imputation for missing beats to generate significantly more realistic ECG waveforms. Upon comparison with SSSD$^{S4}$, PulseDiff-SSSD$^{S4}$ reconstructs a larger amount of ECG waveform and achieves improvements in estimating the beat amplitudes. Note also how the pulse prior has enabled the confidence intervals of the imputed values to become more informed and sensibly bounded. Reconstructed waveforms are also shown for the current SOTA transformer model, BDC Trans, and a standard transformer, Vanilla Trans. Empirically, PulseDiff variants reconstruct more realistic pulse waveforms compared to these deterministic models.

\vspace{1em}
\paragraph{Downstream tasks}
Performance on the classification task (diagnosis, form and rhythm) and on the beat detection task (F1, precision, sensitivity) for 30\% transient and extended missingness are also shown in Tables \ref{tab:results-transient-ptbxl} and \ref{tab:results-extended-ptbxl}, respectively. From Table \ref{tab:results-transient-ptbxl}, we observe that our method improves the baseline diffusion models CSDI on every evaluation metric, and SSSD on most metrics. Moreover, our PulseDiff-SSSD$^{S4}$ distinctively improves the performance of other probabilistic and deterministic models on 5/6 metrics. From Table \ref{tab:results-extended-ptbxl}, we observe that our method improves CSDI by a considerable margin and achieves similar performance with SSSD. However, unlike transient missingness, the advantage of diffusion models in extended missing is not clear. Deterministic models especially BDC still hold the best performance. One possible reason is that diffusion models are trained to estimate the Gaussian noise instead of directly capturing the dependency in data sequence. Therefore, in extended missingness, even if strong guidance is provided, the generated data may have variances compared to the ground-truth sample, leading to results inferior to deterministic models that are optimized by a data-domain loss. The improvement of diffusion models in this setting will be explored in our future work.  

Figure \ref{fig:classification_performance} displays the performance of the diffusion probabilistic models on the downstream tasks for different missingness percentages. From this, PulseDiff-CSDI demonstrates consistent improvements upon CSDI in both transient and extended missingness. In addition, PulseDiff-SSSD$^{S4}$ demonstrates  improvements upon SSSD$^{S4}$ in transient missingness, and performs comparably for extended missingness.

\vspace{1em}
\subsection{Ablation Study}
Table \ref{tab:ablation_results} shows the results of the ablation study conducted on each proposed component in PulseDiff, evaluated for PulseDiff-CSDI with the same hyperparameter setting identified earlier. In the table, `Score' denotes the rhythm confidence score. Validating earlier results, we find that MSE is lower for PulseDiff-CSDI compared to CSDI, and the performance on downstream tasks is higher. Upon comparison of `CSDI + 1 fixed prior' and `CSDI + 1 fixed prior + score', it is observed that introducing the rhythm confidence score lowers MSE and improves performance on downstream tasks. Additionally, by comparing `CSDI + 1 fixed prior' to `CSDI + 16 augmented prior', we find that introducing template augmentation does indeed lower MSE and improve performance on downstream tasks. Therefore, these two ablations validate the utility of the rhythm confidence score and template augmentation components in PulseDiff. However, we observe that the performance of PulseDiff-CSDI is lower compared to `CSDI + 1 fixed prior + score' and `CSDI + 16 augmented prior', suggesting that there is further room to optimise PulseDiff by considering how our two proposed improving techniques are used jointly. One potential method could be to use classifier-free guidance \cite{ho2022classifierfree} to attain a trade-off between template augmentation and the rhythm confidence score to combine their benefits; which will be explored in future work.

\section{Conclusion and Future Directions}
In this work, we introduce PulseDiff, a methodology designed to aid missing value imputation using DDPMs in ECG data. By analyzing the quasiperiodic structure of pulse data, we leverage observed components of the signal to extract a personalized template and compute a template-based prior for the missing values. We also consider the time-varying pulse morphologies by augmenting our prior with random shift terms on the template across both positional and amplitude dimensions. Moreover, we account for the variation of pulse morphologies across different populations by extracting subject-level templates and introducing a confidence score to prevent providing misleading prior information during imputation. The advantages of these improving techniques are carefully studied.

From the perspective of methodology, PulseDiff guides the diffusion models with a template prior which highlights the location and morphology of pulsative signals. It is partly similar to the image generation guided by user-defined sketch. However, the difference is that our manually estimated prior may not be consistent with the ground-truth data, especially when we suffer lots of missing values, e.g., the extended missingness. Therefore, in the future we will study following three topics: $1$) we will improve both diffusion models and our template in the extended missing task; $2$) optimise PulseDiff performance by considering how template augmentation and rhythm confidence score are used jointly; and $3$) extend our methods on more pulsative data such as PPG and blood pressure, to provide a strong generative model based imputation model for this community. Our method has the potential to improve the quality of mHealth ECG recordings and thus improve the ability of automated diagnostic algorithms to provide accurate classification of abnormal ECGs recorded using mHealth technologies.

\section*{Acknowledgments}
Alexander Jenkins is supported by the UKRI CDT in AI for Healthcare \url{http://ai4health.io} (Grant No. P/S023283/1). Fu Siong Ng is supported by the British Heart Foundation (RG/F/22/110078, FS/CRTF/21/24183 and RE/19/4/34215) and the National Institute for Health Research Imperial Biomedical Research Centre.

% {\appendix[Proof of the Zonklar Equations]
% Use $\backslash${\tt{appendix}} if you have a single appendix:
% Do not use $\backslash${\tt{section}} anymore after $\backslash${\tt{appendix}}, only $\backslash${\tt{section*}}.
% If you have multiple appendixes use $\backslash${\tt{appendices}} then use $\backslash${\tt{section}} to start each appendix.
% You must declare a $\backslash${\tt{section}} before using any $\backslash${\tt{subsection}} or using $\backslash${\tt{label}} ($\backslash${\tt{appendices}} by itself
%  starts a section numbered zero.)}


\bibliographystyle{IEEEtran}
\bibliography{bibi}
\end{document}